%% file: 0936.tex
\documentclass[runningheads]{llncs}
\usepackage{graphicx}
\usepackage{amsmath,amssymb} 
\usepackage{color}
\usepackage{mathtools}
\usepackage{subfigure}
\def\eg{e.g.}

\begin{document}
\title{DPP-Net: Device-aware Progressive Search for Pareto-optimal Neural Architectures} 

\titlerunning{DPP-Net}
%
\author{Jin-Dong Dong\inst{1} \and An-Chieh Cheng\inst{1} \and Da-Cheng Juan\inst{2} \and Wei Wei\inst{2} \and Min Sun\inst{1}}
%
\authorrunning{J.-D. Dong, A.-C. Cheng, D.-C. Juan, W. Wei,  M. Sun}
%

\institute{National Tsing-Hua University, Hsinchu, Taiwan \and
Google, Mountain View, CA, USA\\
\email{mark840205@gmail.com, \{anjiezheng@gapp,sunmin@ee\}.nthu.edu.tw, \{dacheng,wewei\}@google.com}}
\maketitle              
\input{Abstract}
\input{Intro}
\input{Related}
\input{Approach}
\input{Experiments}
\input{Conclusion}
\section{Acknowledgement}
We are grateful to the National Center for High-performance Computing for computer time and facilities, and Google Research, MediaTek, MOST 107-2634-F-007-007 for their support. This research is also supported in part by the Ministry of Science and Technology of Taiwan (MOST 107-2633-E-002-001), National Taiwan University, Intel Corporation, and Delta Electronics.

%
%
%
%

\end{document}

%% file: Abstract.tex
\begin{abstract}
Recent breakthroughs in Neural Architectural Search (NAS) have achieved state-of-the-art performances in applications such as image classification and language modeling. However, these techniques typically ignore device-related objectives such as inference time, memory usage, and power consumption. Optimizing neural architecture for device-related objectives is immensely crucial for deploying deep networks on portable devices with limited computing resources. We propose DPP-Net: Device-aware Progressive Search for Pareto-optimal Neural Architectures, optimizing for both device-related (e.g., inference time and memory usage) and device-agnostic (e.g., accuracy and model size) objectives.
DPP-Net employs a compact search space inspired by current state-of-the-art mobile CNNs, and further improves search efficiency by adopting progressive search (Liu et al. 2017).
Experimental results on CIFAR-10 are poised to demonstrate the effectiveness of Pareto-optimal networks found by DPP-Net, for three different devices: (1) a workstation with Titan X GPU, (2) NVIDIA Jetson TX1 embedded system, and (3) mobile phone with ARM Cortex-A53. Compared to CondenseNet and NASNet (Mobile), DPP-Net achieves better performances: higher accuracy \& shorter inference time on various devices. Additional experimental results show that models found by DPP-Net also achieve considerably-good performance on ImageNet as well.

\keywords{Architecture Search, Multi-objective Optimization}
\end{abstract}

%% file: Intro.tex
\section{Introduction}
Deep Neural Networks (DNNs) have demonstrated impressive performance on many machine-learning tasks such as image recognition \cite{krizhevsky2012imagenet}, speech recognition \cite{hannun2014deep}, and language modeling \cite{sutskever2014sequence}. Despite the great successes achieved by DNNs, crafting neural architectures is usually a manual, time-consuming process that requires profound domain knowledge. Recently, automating neural architecture search (NAS) has drawn lots of attention from both industry and academia \cite{negrinho2017deeparchitect,zoph2016neural}. Approaches for NAS can mainly be categorized into two branches: based on Reinforcement Learning (RL)  \cite{pham2018efficient,zoph2016neural,baker2016designing,zoph2017learning,zhong2017practical} or Genetic Algorithm (GA) \cite{real2017large,xie2017genetic,liu2017hierarchical,real2018regularized}. There are also works not based on RL or GA, such as \cite{liu2017progressive}, achieving comparable performance by using other efficient search algorithms. However, most of these works mentioned above focus on optimizing one single objective (\eg, accuracy), and other objectives have been largely ignored, especially thoses related to devices (\eg, latency).

On the other hand, while designing complex, sophisticated architectures have already been treated more like an art than science, searching for neural architectures optimized for multiple objectives has posed an even more significant challenge. To this end, new architectures leveraging novel operations \cite{howard2017mobilenets,zhang2017shufflenet,huang2017condensenet} have been developed to achieve higher computing efficiency than conventional convolution. Not surprisingly, designing such architectures requires, again, profound domain knowledge and much effort. Therefore, how to automatically search for network architectures jointly considering high accuracy and other objectives (\eg, inference time, model size, etc. to conforms to device-related constraints) remains a critical yet less addressed question. To the best of our knowledge, there is one previous work \cite{kimnemo} that searches network architectures by considering both accuracy and inference time. Nevertheless, the computational power required during training by their algorithm is very significant, and their search space is naively small.

\begin{figure}[t!]
\begin{minipage}{0.55\textwidth} 
\includegraphics[width=\textwidth]{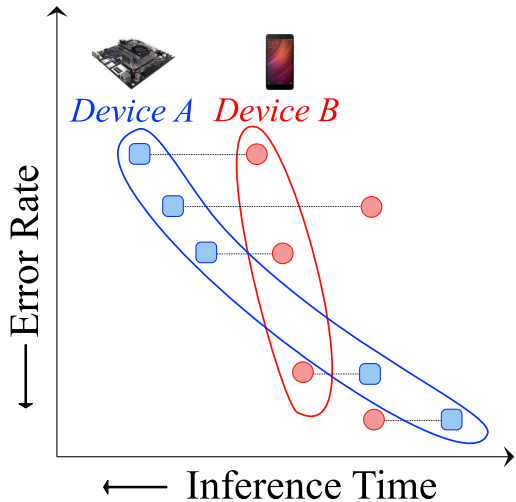}
\end{minipage}%
\hspace{.01\textwidth}
\begin{minipage}{0.4\textwidth}
\caption{Different devices share different Pareto-Optimality. An optimal point on Device A's Pareto front may not lie on Device B's Pareto front. Given multiple device-related (\eg, inference time and memory usage) and device-agnostic (\eg, accuracy and model size) objectives. Our DPP-Net can efficiently find various network architectures at the Pareto-front for the corresponding device.}
\end{minipage}
\label{fig.concept}
\end{figure}

We propose \textit{DPP-Net}: \textit{D}evice-aware \textit{P}rogressive Search for \textit{P}areto-optimal Neural Architectures given multiple device-related (\eg, inference time and memory usage) and device-agnostic (\eg, accuracy and model size) objectives. It is an efficient search algorithm to find various network architectures at the Pareto-front (Fig.\ref{fig.concept}) in the multiple objectives space to explore the trade-off among these objectives. In this way, a deep learning practitioner can select the best architecture under the specific use case.
We define our search space by taking inspirations from state-of-the-art handcrafted mobile CNNs, which is more compact and efficient comparing to usual NAS architectures. For search efficiency, we have also adopted the progressive search strategy used in \cite{liu2017progressive} to speed up the search process. Experimental results on CIFAR-10 demonstrate that DPP-Net can find various Pareto-optimal networks on three devices: (1) a workstation with Titan X GPU, (2) NVIDIA Jetson TX1 embedded system, and (3) a mobile phone with ARM Cortex-A53. Most importantly, DPP-Net achieves better performances in both (a) higher accuracy and (b) shorter inference time, comparing to the state-of-the-art CondenseNet on three devices. Finally, our searched DPP-Net achieves considerably good performance on ImageNet as well.

%% file: Related.tex
\section{Related Work}
Recent advancements on neural architecture search can be classified into three basic categories: Reinforcement Learning (RL) based approaches,  Genetic Algorithm (GA) based ones and the third category of methods that involve optimization techniques other than those two. In addition to architecture search techniques, we will also focus on those methods that work on multiple objectives.

\paragraph{RL-based approach.}
Seminal work by \cite{zoph2016neural} proposed ``Neural Architecture Search (NAS)'' using REINFORCE algorithm \cite{Williams:1992:SSG:139611.139614} to learn a network architecture called ``controller'' RNN that generates a sequence of actions representing the architecture of a CNN. Classification accuracies of the generated bypass CNN models on a validation dataset are used as rewards for the controller. NASNet \cite{zoph2017learning} further improves NAS by replacing REINFORCE with proximal policy-optimization (PPO) \cite{schulman2017proximal} and search architectures of a ``block'' which repeatedly concatenated itself to form a complete model. This techniques has not only reduced the search space but also managed to incorporate empirical knowledge when designing a CNN. Other works in the field including approach used in \cite{cai2018efficient} which searches model architectures by manipulating the depth of the width of the layers using policy gradient, and the methods proposed by \cite{baker2016designing} and \cite{zhong2017practical} which search network architectures using Q-learning. A concurrent work \cite{pham2018efficient} proposed a model to force all child networks to share weights, which largely reduced the computational costs needed to search in a space as defined by \cite{zoph2017learning}.

\paragraph{GA-based approach.}
Except for RL-based methods, Genetic Algorithm based methods \cite{real2017large,xie2017genetic,liu2017hierarchical} are also popular in architecture search research. One of the recent work in this field \cite{real2018regularized} achieves state-of-the-art performance on CIFAR-10 image classification task over RL-based method.

\paragraph{Other approaches}
Methods using either RL-based or GA-based algorithm usually requires a significant amount of computational power and are therefore infeasible in certain situations. Many approaches are proposed to specifically address this issue by proposing their search strategies that cannot be categories using methods that belong to the previous two families. \cite{negrinho2017deeparchitect} use Monte Carlo Tree Search (MCTS) to search through the space of CNN architectures in a shallow-to-deep manner and randomly select which branch to expand at each node. A sequential Model-based Optimization (SMBO) \cite{hutter2011sequential} that learns a predictive model is further adopted to help the decision making of node expansion. \cite{liu2017progressive} also use SMBO as the search algorithm and have achieved comparable performance to NASNet using significantly less computational resources while operated on the same search space. \cite{baker2018accelerating} proposed to predict performance to reduce the effort of searching model architectures. A concurrent work \cite{brock2017smash} proposed to train a network to predict the weights of another network and combine this method with random search to search for good candidate models. Despite the small number of resources required in each search, the performance of the model is hard to compete with state-of-the-art approaches.

\paragraph{Architecture search with multiple objectives.}
All the previously mentioned works focus on searching models that achieve highest performance (\eg classification accuracy) regardless of the model complexity. \cite{kimnemo} proposed to treat neural network architecture search as a multi-objective optimization task and adopt an evolutionary algorithm to search models with two objectives, run-time speed, and classification accuracy. However, the performances of the searched models are not comparable to handcrafted small CNNs, and the numbers of GPUs they required are enormous.

\paragraph{Handcrafted models with multiple objectives.}
The machine learning and computer vision community are rich in handcrafted neural architectures. Here we will list some of the most recent work that involves multiple objectives. \cite{howard2017mobilenets} and ShuffleNet \cite{zhang2017shufflenet} have utilized depth-wise convolution and largely reduced computational resources required but remained comparably accurate. However, the real-world implementation of depth-wise convolution in most of the deep learning framework have not reached the theoretical efficiency and results in much inferior inference time. CondenseNet \cite{huang2017condensenet} proposed to use a group convolution \cite{krizhevsky2012imagenet} variant in order to achieve state-of-the-art computational efficiency.

%% file: Approach.tex
\section{Search Architecture}
\begin{figure}[h]
\centering     
\includegraphics[width=0.85\textwidth]{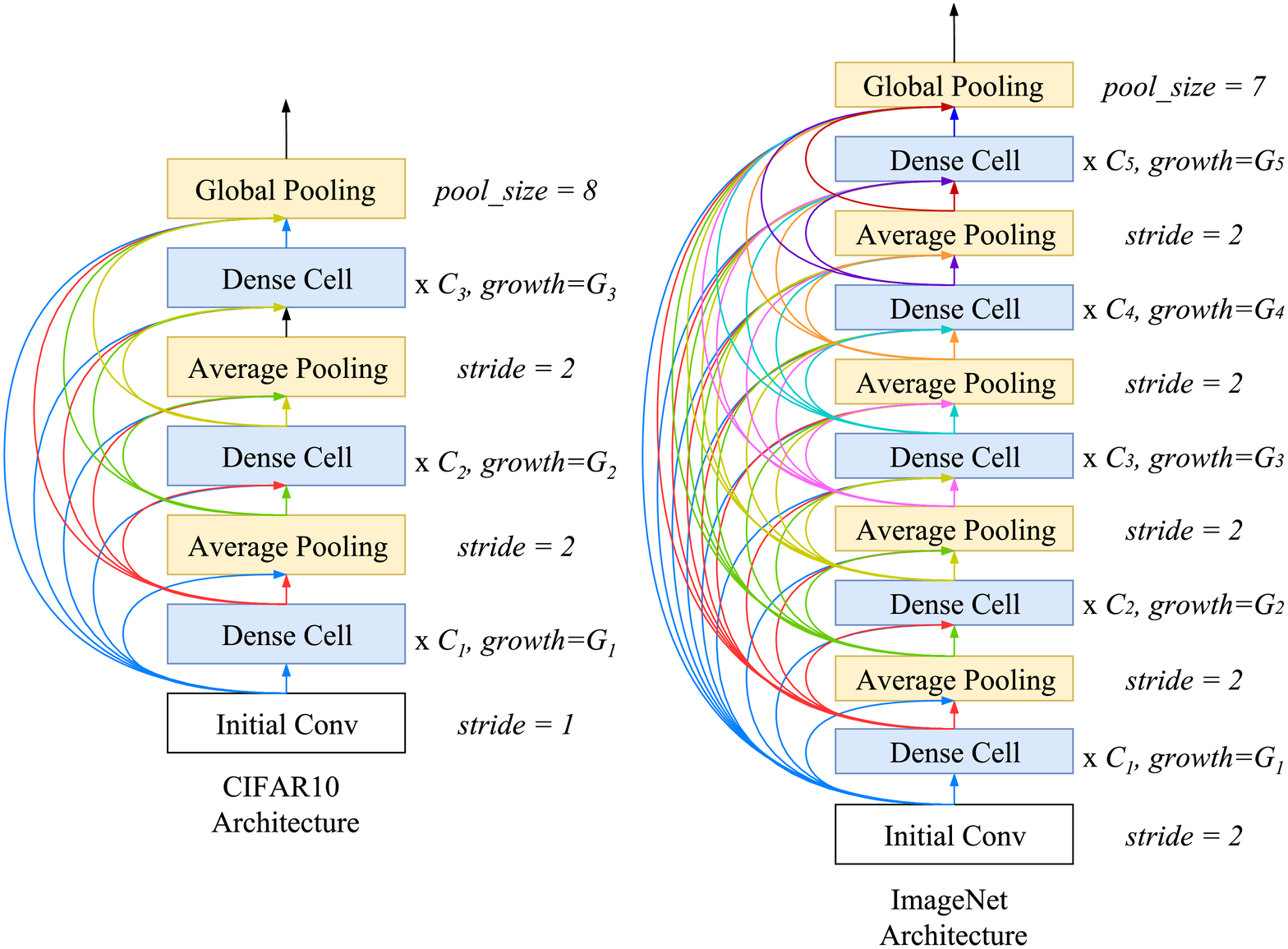}
\caption{\textbf{Network architecture for CIFAR-10 and ImageNet.}
Our final network structure is fully specified by defining the Dense Cell topology. The number of Dense Cell repetitions $C$ and the growth rate $G$ are different for CIFAR-10 and ImageNet architecture. Note that in ImageNet Architecture we set the stride value of initial convolution 2 and the pool size in global pooling to 7 due to the scale of the input image.}
\label{fig.bigdiagram}
\end{figure}

In Fig.\ref{fig.bigdiagram} we illustrate the overall architectures. We repeat an identical ``cell'' (Dense Cell) numerous of times following the connecting rules of CondenseNet \cite{huang2017condensenet}. We take inspirations from CondenseNet, which optimizes both classification accuracy and inference speed for mobile devices. The feature maps are directly connected even with different resolution and the growth rate is doubled whenever the size of the feature maps reduces. The strategy to make fully dense connections encourage feature re-use and the exponentially increased growth rate reduces computational costs. These characteristics are beneficial when deploying models on energy constrained devices. As we conduct our searching on CIFAR-10, transferring the searched model to ImageNet, requires more stride 2 pooling layers and Dense Cells since the size of the input images (224 x 224) is way larger than CIFAR10 (32 x 32). Finally, a global average pooling layer is appended to the last Dense Cell to obtain the final output.

The overall architectures (\eg, how many cells are connected, initial output feature map size, growth rate) are fixed before searching, the only component we are going to search is the cell structure, this idea follows the heuristics of searching for a ``block'' similar to \cite{zoph2017learning,liu2017progressive}. Each cell to be searched consists of multiple layers of two types - normalization (Norm) and convolutional (Conv) layers. We progressively add layers following the Norm-Conv-Norm-Conv order (Fig.\ref{fig.searchspace}(a)-Right). The operations available for Norm (yellow boxes) and Conv (green boxes) layers are shown in the left and right column below, respectively:

\begin{multicols}{2}
\begin{enumerate}
	\item Batch Normalization + Relu
	\item Batch Normalization
    \item No op (Identity)
\end{enumerate}
\hfill\linebreak
\begin{enumerate}
	\item 1x1 Convolution
	\item 3x3 Convolution
    \item 1x1 Group Convolution
    \item 3x3 Group Convolution
    \item 1x1 Learned Group Convolution
    \item 3x3 Depth-wise Convolution
\end{enumerate}
\end{multicols}

\section{Search Space}
\begin{figure}[h]
\begin{center}
\includegraphics[width=0.95\linewidth]{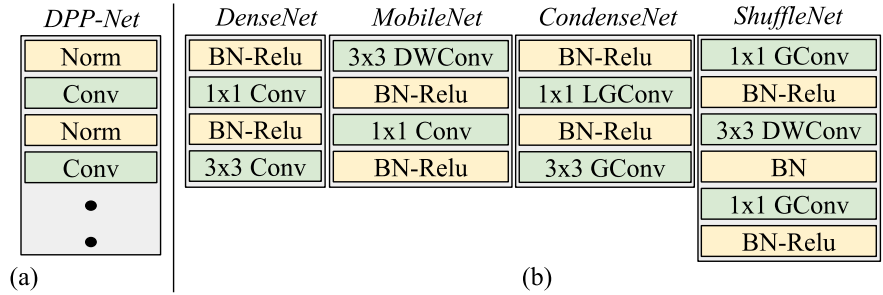}
\end{center}
\caption{\textbf{Search Space Design.}
Panel (a):  We show the cell structure of our DPP-Net. Panel (b): cells of efficient CNNs. BN, DW, LG, G stands for Batch Norm, Depth-wise, Learned Group, Group, respectively. All the group convolutions are implicitly followed by channel shuffle operation.}
\label{fig.searchspace}
\end{figure}

Our search space covers well-designed efficient operations (\eg, Depth-wise Convolution \cite{chollet2016xception}, Learned Group Convolution \cite{huang2017condensenet}) to take advantages of empirical knowledge when designing efficient CNNs. This not only ensures the robustness and efficiency of our searched architectures but also reduces the training time of the searched model, therefore reduce the search time as well. Finally, the block of other efficient CNNs, \eg, MobileNet \cite{howard2017mobilenets}, ShuffleNet \cite{zhang2017shufflenet} are also shown in Fig.\ref{fig.searchspace}(b) for a more thorough comparison.

We now measure the complexity of our search space to have an intuition of the size of the search problem. For a $\ell$-layer cell, the total number of possible cell structures is $O_{0} \times O_{1} \times ... \times O_{i} \times ... \times O_{\ell} \text{ where } O_{i} = \left | Norm \right | \text{ if } i \text{ mod } 2 = 0 \text{ or } O_{i} = \left | Conv \right |$. As shown above, the number of operations in the Norm set is 3 and the number of operations in the Conv set is 6. Therefore, a 3-layer cell structure has $3 \times 6 \times 3 = 54$ possibilities, a 4-layer cell will have $54 \times 6 = 324$ possible structures. As the number of layer increases, it is hardly pragmatic to train all the architectures. This search space is undersized comparing to the search space of \cite{zoph2017learning,liu2017progressive} because we discarded the operations that are rarely used in modern mobile CNNs and we do not need to search for which layer to connect to. Nevertheless, this search space is still versatile enough to cover a wide variety of possible mobile models.

\section{Search Algorithm}
\begin{figure}[h]
\begin{center}
\includegraphics[width=0.95\linewidth]{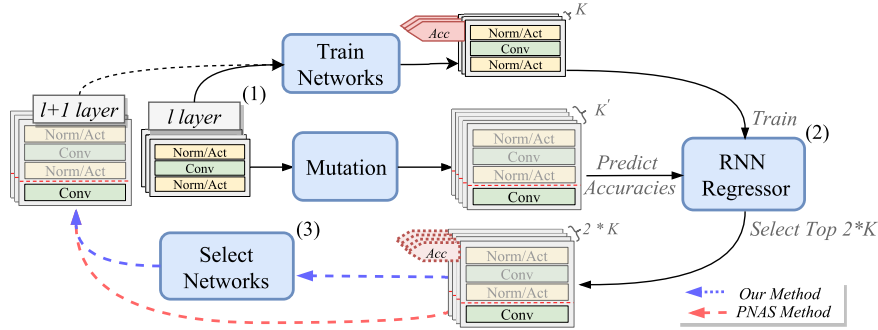}
\end{center}
\caption{\textbf{Flow Diagram of Our Search Algorithm}.
We adopt Sequential Model-Based Optimization (\cite{hutter2011sequential}) algorithm to search efficiently with the following three steps: \textbf{(1)Train and Mutation}, \textbf{(2)Update and Inference}, and \textbf{(3)Model Selection}. Note that $\ell$ is the layers in a cell, $K$ is the number of models to train, and ${K}'$ is the number of models after \textit{Mutation}.}
\label{fig.diagram}
\end{figure}

\subsection{Overview}
Many architecture search approaches intuitively search on the complete search space which requires significant computing power. Inspired by \cite{liu2017progressive}, which progressively search the architectures from a small search space to a large one, we adopt Sequential Model-Based Optimization (\cite{hutter2011sequential}) algorithm to navigate through the search space efficiently. Our search algorithm consists of the following three main steps (Fig.~\ref{fig.diagram}).

\begin{enumerate}
\item{\textbf{Train and Mutation}.} In this stage, we train $K$ $\ell$-layer models and acquire their accuracies after $N$ epochs. Meanwhile, for each $\ell$-layer model, we mutate it and acquire a $\ell+1$-layer model by exploring all possible combinations. Assuming that we have $K$ models before mutation, the number of models after mutation $K'$ process will be the following.

\begin{equation}\label{eq.mutation}
  {K}' =
  \begin{dcases}
  K \times \left | Norm \right |, & \text{if } \ell + 1 \text{ mod } 2 = 0 \\
  K \times \left | Conv \right |, & \text{otherwise}
  \end{dcases}
\end{equation}

\item{\textbf{Update and Inference}.} In \textbf{Train and Mutation} step, the algorithm will generate a large number of candidate models that are usually beyond our ability to evaluate. We use a surrogate function to predict the networks' accuracies with the given architectures. The surrogate function is updated with the evaluation accuracies (output) and the architectures (inputs) of the $K$ $\ell$-layer models from the \textbf{Train and Mutation} step. After the surrogate function is updated, we predict the accuracies of the mutated $\ell+1$-layer models. Using a surrogate function avoids time-consuming training to obtain true accuracy of a network with only a slight drawback of regression error.

\item{\textbf{Model Selection}.} There are two ways we can select $\ell+1$-layers models. \\
\textit{PNAS Method.} \cite{liu2017progressive} adopted the SMBO algorithm to search for block architectures of increasing complexity. During the search process, SMBO simply selects top $K$ performing models based on predicted accuracies. This approach is inconsiderate of the heterogeneity of real-world portable device, which is only equipped with limited power supply.

\textit{Our Method.} Our method considers not only the accuracy of the models but also the device-aware characteristics. Those characteristics include QoS (Quality of Service) and hardware requirements (e.g., memory size), which are critical metrics to be considered on mobile and embedded devices. Given the device we are searching on, multiple hard constraints $\mu$ and soft constraints $\xi$ are set. A hard constraint $\mu$ is considered to be the minimal requirement of the model. A model that does not meet the hard constraint will be removed from the candidate list. On the other hand, a soft constraint $\xi$ is treated as one of the objectives to be optimized which will be eventually selected using Pareto Optimality selection.

\end{enumerate}

\begin{figure}[h!]
\begin{center}
\includegraphics[width=0.9\linewidth]{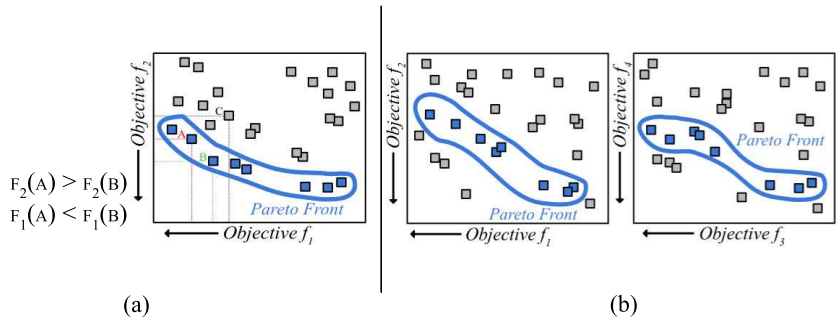}
\end{center}
\caption{\textbf{A symbolic figure for Pareto Optimality.}
Panel (a) illustrates an example of two objectives Pareto front. Every box represents a feasible choice. In this case, our goal is to minimize both objectives. Since box C is dominated by both box A and box B, it is not on the Pareto front. While box A and box B both lie on the front because none of them dominated another. Panel (b) demonstrates that when the number of objectives is more than two, the Pareto front becomes more complicated.}
\label{fig.pareto}
\end{figure}

\subsection{Pareto Optimality}
Since we are optimizing the problem using multiple objectives, no single solution will optimize each objective simultaneously and compromises will have to be made. We treat neural network architecture search as a multi-objective optimization problem and use Pareto Optimality over a set of pre-defined objectives to select models. Using Pareto Optimization, it is likely that there exists a number of optimal solutions. A solution is said to be Pareto optimal if none of the objectives can be improved without worsening some of the other objectives, and the solutions achieve Pareto-optimality are said to be in the Pareto front. 

\subsection{Surrogate Function}
To accurately predict the classification accuracy of an architecture, a surrogate function is used. The surrogate function is able to learn efficiently from a few data points and handle variable-sized inputs (models with different number of layers). Hence, we choose a Recurrent Neural Network (RNN), the last hidden state of the RNN is followed by a fully connected layer with sigmoid nonlinearity to regress accuracy. The reason for choosing RNN as the surrogate function is because of its high sampling efficiency and the ability to handle different length of inputs. The input to the RNN is the one-hot encoding of our cell structure and each structure has its own embedding.

\begin{figure}[h]
\begin{minipage}{0.5\textwidth} 
\includegraphics[width=\textwidth]{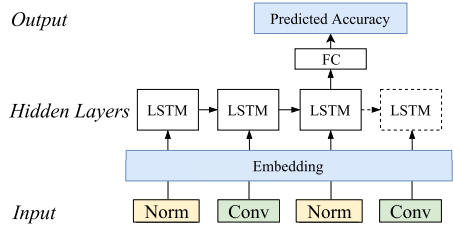}\label{fig.surrogate}
\end{minipage}%
\hspace{.01\textwidth}
\begin{minipage}{0.4\textwidth}
\caption{The architecture diagram of our Recurrent Neural Network (RNN). The dashed block indicates we progressively search for more layers architectures.}
\end{minipage}
\end{figure}

%% file: Experiments.tex
\section{Experiments and Results}

\subsection{Experimental Details}
We conduct our search on the CIFAR-10 dataset with standard augmentation, the training set consists of 50,000 images and the testing set consists of 10,000 images. After the search is done, we use the cell structure to form a larger model and train on ImageNet \cite{deng2009imagenet} classification task to see how well the search performs. For the surrogate function, we use a standard LSTM with layer normalization \cite{ba2016layer}, the hidden state size and the embedding size are both set to 128. Bias in the fully connected layer is initialized to 2, and the embeddings use random uniform initializer in range 0 to 1. To train the surrogate function, we use Adam Optimizer \cite{kingma2014adam} with learning rate 0.008.

During the search, the number of repeated blocks $C_1, C_2, C_3$ are set to 14, 14, 14, $G_1, G_2, G_3$ are set to 8, 16, 32 for CIFAR-10 and the searching end at $\ell$ = 4. Each sampled architecture is trained for 10 epochs with batch size 256 using Stochastic Gradient Descent and Nesterov momentum weight 0.9. The learning rate is set to 0.1 with cosine decay \cite{loshchilov2016sgdr}. At each iteration of the search algorithm, our number of models to train, $K$, is set to 128. After searching is done, we train the final models on ImageNet with batch size 256 for 120 epochs, the number of repeated blocks $C_1, C_2, C_3, C_4, C_5$ are set to 4, 6, 8, 10, 8 and $G_1, G_2, G_3, G_4, G_5$ are set to 8, 16, 32, 64, 128.

The detail settings of the devices to search on are shown in Table.\ref{tb.hwsettings}. When searching models on WS and ES, we consider 4 objectives, evaluation error rate, number of parameters, FLOPs, and actual inference time on different computing devices. While on Mobile Phone, we consider an additional metric, memory usage, as our $5^{th}$ objective.

\begin{table}[h]
\caption{\textbf{Hardware Specifications and Numbers of Objectives.}
For WS, 64 GB is the CPU memory and 12 GB is the GPU memory. In ES, memory space is shared among CPU and GPU}
\label{tb.hwsettings}
\centering
\resizebox{1\columnwidth}{!}{
\begin{tabular}{|c|c|c|c|}
\hline
           & \textbf{Workstation (WS)} & \textbf{Embedded System (ES)} & \textbf{Mobile Phone (M)} \\ \hline
Instance   & Desktop PC           & NVIDIA Jetson TX1  & Xiaomi Redmi Note 4           \\ \hline           
CPU        & Intel i5-7600        & ARM Cortex57      & ARM Cortex53               \\ \hline
Cores      & 4                    & 4                   & 8                            \\ \hline
GHz        & 3.5                  & 1.9                 & 2.0                          \\ \hline
CUDA       & Titan X (Pascal)     & Maxwell 256         & -                            \\ \hline
Memory     & 64 GB / 12 GB        & 4 GB                & 3 GB                         \\ \hline
Objectives & 4                    & 4                   & 5                            \\ \hline
\end{tabular}}
\end{table}

\subsection{Results on CIFAR-10}
We first provide the results about the Pareto-optimal candidates (each trained for 10 epochs) found during the search process, and then demonstrate the evaluations of final models (trained for 300 epochs).

\begin{figure}[h]
\centering
\subfigure[]{\label{fig.params}\includegraphics[width=0.325\textwidth]{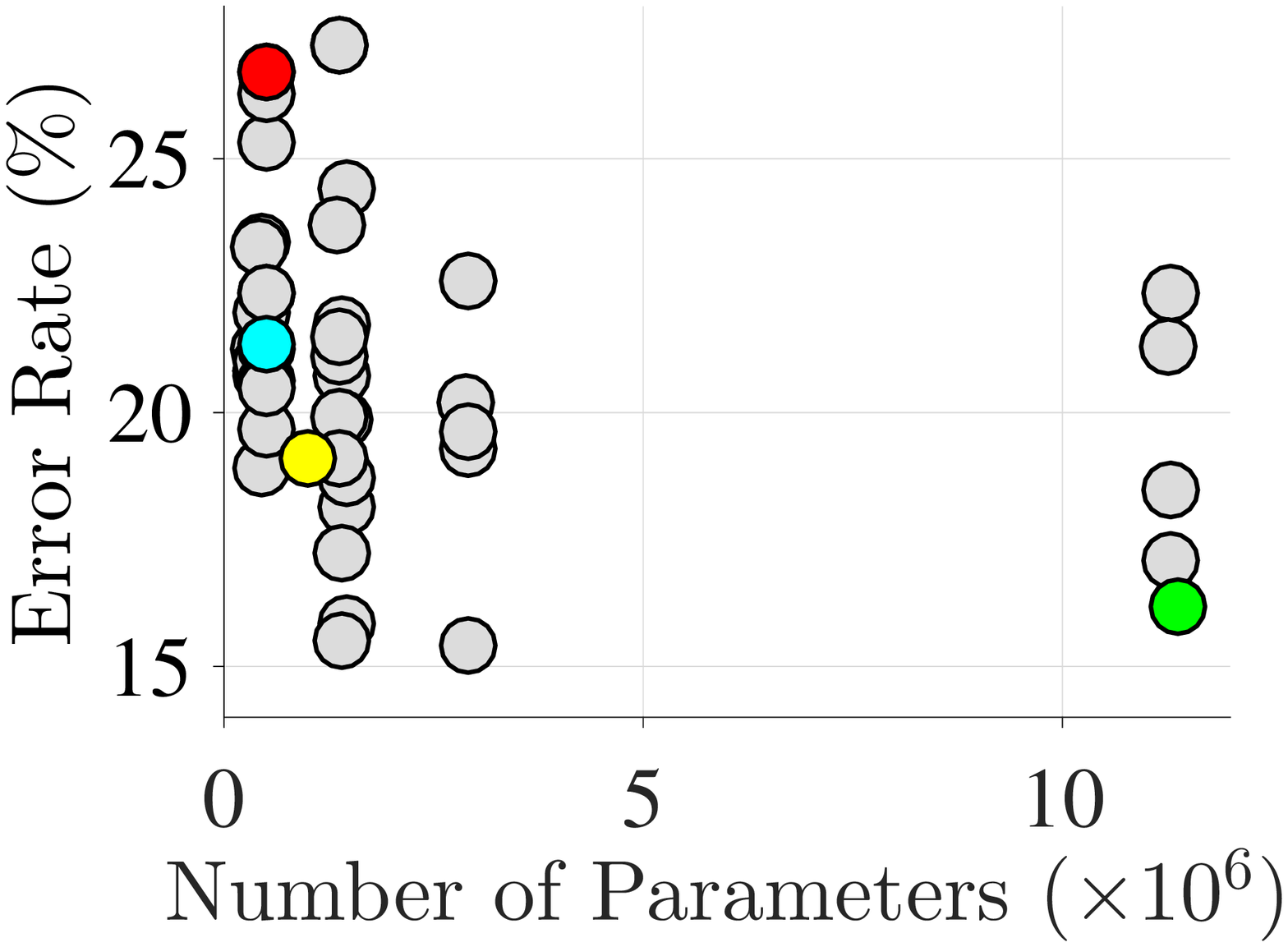}}
\subfigure[]{\label{fig.flops}\includegraphics[width=0.325\textwidth]{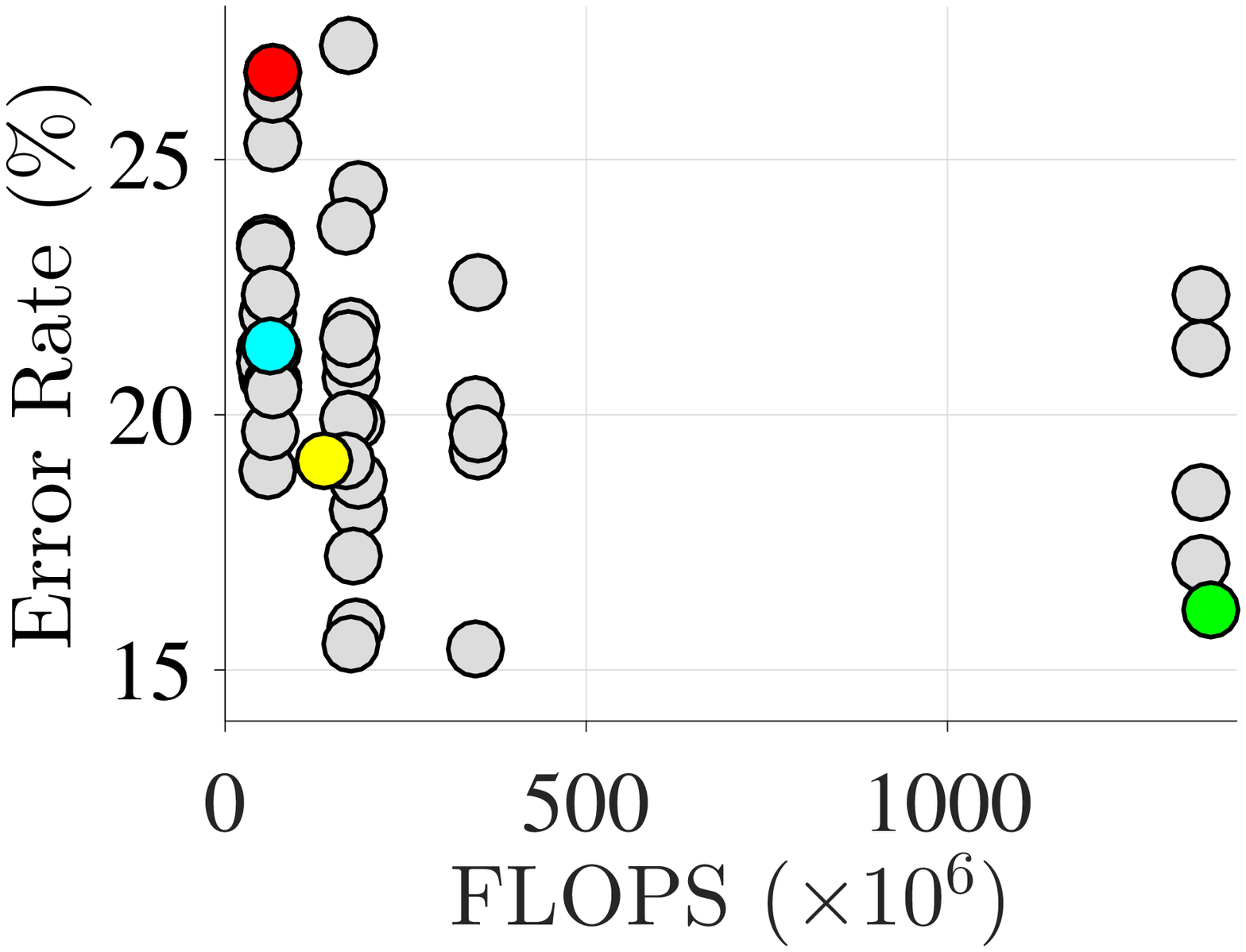}}
\subfigure[]{\label{fig.dur}\includegraphics[width=0.325\textwidth]{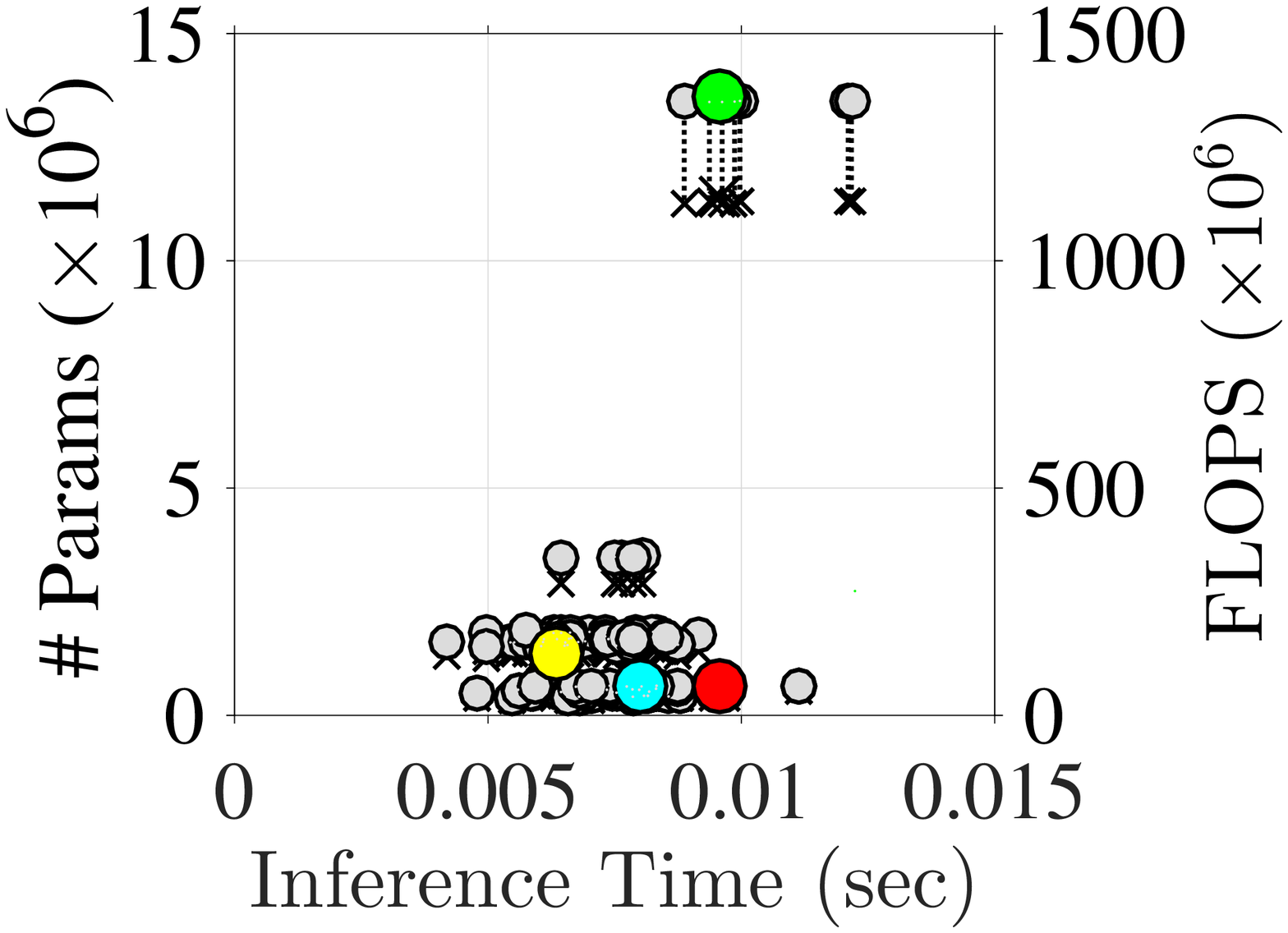}}
\caption{\textbf{Pareto-optimal candidates on WS (trained with 10 epochs) evaluated with Cifar10 dataset}.
(a) is the scatter plot between error rate (Y-axis) and the number of parameters (X-axis), whereas (b) stands for error rate v.s. FLOPs. (c) is the number of parameters (left Y-axis) and FLOPs (right Y-axis) v.s. actual inference time (X-axis), where the dot represents params v.s. inference time and the cross is FLOPs v.s. inference time. Each model is color-coded: green (DPP-Net-PNAS), yellow (DPP-Net-WS), and cyan (DPP-Net-Panacea). Notice that each candidate here represents a neural architecture that achieves Pareto optimality. Finally, CondenseNet (red dots) is included for comparison.}
\label{fig.paretofront}
\end{figure}

Fig.\ref{fig.paretofront} shows the candidates extracted from the Pareto front during the search process. In Fig.\ref{fig.paretofront}(a,b), no clear pattern (or association) is observed between the error rate and the number of parameters (or FLOPs). Similarly, from Fig.\ref{fig.paretofront}(c), the inference time couldn't be simply associated with the device-agnostic objectives (FLOPs and number of parameters). As we will show later in Table.\ref{tb.cifar} and Fig.\ref{fig.sort_time}, not surprisingly, inference time is device-dependent since, in addition to modeling, the hardware implementation also affects the inference time. For a better comparison and also to showcase our DPP-Net, we evaluate and plot the performance of CondenseNet (reproduce 10 epochs performance), which is also included in our search space but not on the Pareto front.

\begin{table}[t]
\caption{\textbf{Cifar10 Classification Results.}
Missing values are the metrics not reported in their original papers. Pareto front visualizations of our searched networks can also be found in Fig.~\ref{fig.paretofront}. The standard deviation of the metrics of DPP-Net-Panacea are calculated across 10 runs}
\label{tb.cifar}
\centering
\resizebox{1\columnwidth}{!}{
\begin{tabular}{l|lll|llll}
\hline
                                   & \multicolumn{3}{c|}{\textit{Device-agnostic metrics}} & \multicolumn{4}{c}{\textit{Device-aware metrics}}    \\ \hline
\textbf{Model from previous works} & Error rate     & Params      & FLOPs       & Time-WS & Time-ES & Time-M & Mem-M \\ \hline
Real et al. \cite{real2017large}   & 5.4              & 5.4M        & -           & -       & -       & -           & -          \\
NASNet-B \cite{zoph2017learning}   & 3.73             & 2.6M        & -           & -       & -       & -           & -          \\
PNASNet-1 \cite{liu2017progressive}& 4.01             & 1.6M        & -           & -       & -       & -           & -          \\ \hline
DenseNet-BC (k=12) \cite{huang2017densely} & 4.51     & 0.80M       & -           & -       & -       & 0.273       & 79MB       \\
CondenseNet-86 \cite{huang2017condensenet} & 5.0      & 0.52M       & 65.8M       & 0.009   & 0.090   & 0.149       & 113MB      \\ \hline
                                   & \multicolumn{3}{c|}{\textit{Device-agnostic metrics}} & \multicolumn{4}{c}{\textit{Device-aware metrics}}    \\ \hline
\textbf{Model from DPP-Net}        & Error rate       & Params      & FLOPs       & Time-WS & Time-ES & Time-M & Mem-M \\ \hline
DPP-Net-PNAS     & \textbf{4.36}    & 11.39M      & 1364M       & 0.013   & 0.062   & 0.912       & 213MB      \\ \hline
DPP-Net-WS     & 4.78             & 1.00M       & 137M        & \textbf{0.006} & 0.075   & 0.210       & 129MB      \\
DPP-Net-ES     & 4.93             & 2.04M       & 270M        & 0.007   & \textbf{0.044} & 0.381       & 100MB      \\
DPP-Net-M & 5.84             & \textbf{0.45M} & \textbf{59.27M} & 0.008   & 0.065   & \textbf{0.145}  & \textbf{58MB}       \\ \hline
DPP-Net-Panacea  & 4.62 $\pm$ 0.23      & 0.52M       & 63.5M       & 0.009 $\pm$ 7.4e-5  & 0.082 $\pm$ 0.011   & 0.149 $\pm$ 0.017       & 104MB     
\end{tabular}}
\end{table}

During the searching process, the surrogate function was updated several times. The best regression error (on the validation set) is around 12\%. At the first glance, this number is a bit large in terms of predicting the true accuracy. However, it is important to clarify that the purpose of using the surrogate function is to suggest what kind of models may have a relatively good accuracy instead of exactly how accurate the models are. For the search time, we use 4 GTX 1080 GPUs and search for two days (around 48 hours).

\begin{figure}[h]
\begin{center}
\includegraphics[width=1\textwidth]{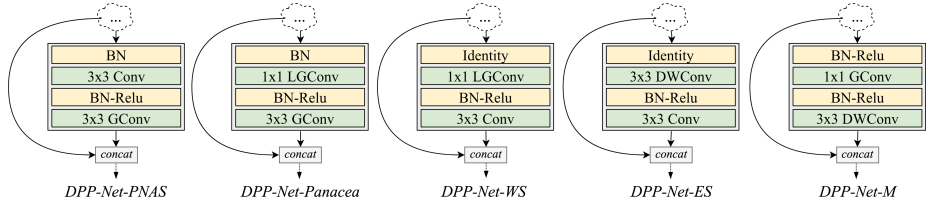}
\end{center}
\caption{\textbf{The result of our searched dense cell topology.}}
\label{fig.mobile_archis}
\end{figure}

After searching process is done, we select two architectures (from others on the Pareto front) for detailed evaluation: DPP-Net-\textit{Device} and DPP-Net-Panacea. DPP-Net-\textit{Device} has a small error rate and the shortest inference time when running on certain \textit{Device} (WS or ES), whereas DPP-Net-Panacea also has a small error rate and performs relatively well on every objective (but longer inference time than DPP-Net-\textit{Device}). These two best models, in terms of Pareto Optimality, are trained for 300 epochs and the evaluation metrics are reported in Table.\ref{tb.cifar} (bottom half). We also include the results of the neural architecture searched by DPP-Net with PNAS \cite{liu2017progressive} criterion: the highest classification accuracy among all the candidates. Furthermore, for the completeness and comprehensive study, in the top half of Table.\ref{tb.cifar}, we include the results from the best models of previous NAS works \cite{zoph2017learning,real2017large,liu2017progressive}, as well as the current state-of-the-art handcrafted mobile CNN models (bottom half) \cite{huang2017condensenet,huang2017densely}. The architectures of these models are shown in Fig.\ref{fig.mobile_archis}.

DPP-Net-PNAS results in finding models with possible large number of parameters and very slow inference time. Our results are compared with state-of-the-art handcrafted mobile CNNs (second group) and models designed using architecture search methods (first group). Our DPP-Net clearly strikes better trade-off among multiple objectives.

\subsection{Results on ImageNet}
We further transfer our searched architecture to test the performance on ImageNet classification task. The cell structures searched using CIFAR-10 dataset are directly used for ImageNet with only a slight modification on the number of repeated Dense Cells. The hyper-parameters for training DPP-Net on ImageNet are nearly identical to training DPP-Net on CIFAR-10, except for the parameter of group lasso regularizer which we set to 1e-5. This regularization induces group-level sparsity for Learned Group Convolution as suggested in \cite{huang2017condensenet}.

The results of ImageNet training is shown in Table.\ref{tb.imagenet}. DPP-Net-Panacea performs better in nearly every aspect than Condensenet-74. Moreover, DPP-Net-Panacea outperforms NASNet (Mobile), a state-of-the-art mobile CNN designed by an architecture search method \cite{zoph2017learning} in every metrics. We further argue that the sophisticated architecture makes NASNet (Mobile) not practical on mobile devices although it has a relatively small number of parameters compared to traditional CNNs. These results again show the versatility and robustness of our device-aware search method.

\begin{table}[h]
\caption{\textbf{ImageNet Classification Results.}
Time-M and Mem-M is the inference time and memory usage of the corresponding model on our mobile phone using ONNX and Caffe2. Due to operations not supported on this framework, we cannot measure the inference time and memory usage of NASNet (Mobile) on our mobile phone}
\label{tb.imagenet}
\centering
\resizebox{1\columnwidth}{!}{
\begin{tabular}{l|lllllll}
\textbf{Model} &\textbf{Top-1} &\textbf{Top-5} &\textbf{Params} &\textbf{FLOPs}  &\textbf{Time-ES} &\textbf{Time-M} &\textbf{Mem-M}  \\ \hline
Densenet-121 \cite{huang2017densely}           & 25.02	 & 7.71 & -	     & -	   & 0.084  & 1.611  & 466MB   \\
Densenet-169 \cite{huang2017densely}           & 23.80   & 6.85 & -	     & -	   & 0.142  & 1.944  & 489MB   \\
Densenet-201 \cite{huang2017densely}           & 22.58	 & 6.34 & -	     & -	   & 0.168  & 2.435  & 528MB   \\ \hline
ShuffleNet 1x (g=8)                            & 32.4    & -    & 5.4M   & 140M    & 0.051  & 0.458  & 243MB   \\
MobileNetV2                                    & 28.3    & -    & 1.6M   & -       & 0.032  & 0.777  & 270MB   \\ 
Condensenet-74 (G=4)\cite{huang2017condensenet}& 26.2	 & 8.30 & 4.8M   & 529M    & 0.072  & 0.694  & 238MB   \\ \hline
NASNet (Mobile)                                & 26.0	 & 8.4	& 5.3M   & 564M    & 0.244  & -      & -       \\ \hline
DPP-Net-PNAS                                   & 24.16   & 7.13 & 77.16M & 9276M   & 0.218  & 5.421  & 708MB   \\ 
DPP-Net-Panacea       & 25.98   & 8.21 & 4.8M   & 523M    & 0.069  & 0.676  & 238MB   \\ 
\end{tabular}}
\end{table}

\subsection{Device Performance Study}
\begin{figure}[h]
\begin{center}
\includegraphics[width=1\linewidth]{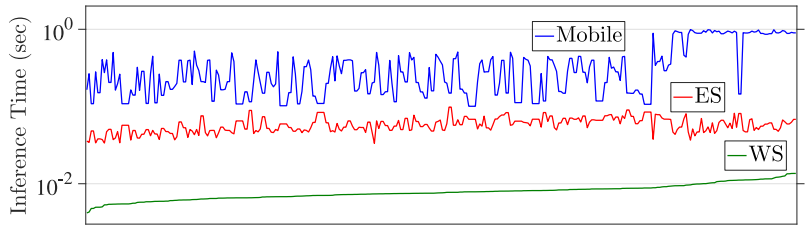}
\end{center}
\caption{\textbf{Models inference time on different devices}.
We show that the inference time is highly device-related. The X-axis is the index of all 4-layers models, sorted by inference time on WS in ascending order.}
\label{fig.sort_time}
\end{figure}

Our main idea is that models searched on one device does not necessarily guarantee good performance on the other devices when it comes to device-related metrics, such as actual inference time. A small number of parameters or FLOPs does not always indicate fast inference time, this is due to existing problems of hardware optimization and software implementations (\eg, implementation of depth-wise convolution is inefficient and group convolution cannot reach theoretical speedups). To prove that inference time is device-aware, we measured the inference time of all 4-layers models (measuring only the network forward time can be done very efficiently) on 3 devices and plot them in Fig.\ref{fig.sort_time}. For WS and ES environments, we test our models on PyTorch 0.3.0 \cite{paszke2017automatic} built with Python 3.5, CUDA-8.0, and CUDNN-6.0, as for M, we follow the instructions from the PyTorch official guide and port the models to Caffe2 for deployment. The X-axis in Fig.\ref{fig.sort_time} is the inference time of all the 4-layer cell structures sorted by WS (green line/bottom line) in ascending order. The red line and the blue line is the inference time on ES and M, respectively.

The plot shows that even on similar devices with identical software settings (WS v.s. ES), the inference time can be sensitive to particular devices. Moreover, inference time on M is significantly disparate to that of WS. Therefore, we conclude that only searching models on an actual device can ensure the robustness of the searched results.

%% file: Conclusion.tex
\section{Conclusions}

Our proposed \textit{DPP-Net} is the first device-aware neural architecture search approach outperforming state-of-the-art handcrafted mobile CNNs.
Experimental results on CIFAR-10 demonstrate the effectiveness of Pareto-optimal networks found by DPP-Net, for three different devices: (1) a workstation with NVIDIA Titan X GPU, (2) NVIDIA Jetson TX1 embedded system, and (3) mobile phone with ARM Cortex-A53. Compared to CondenseNet and NASNet (Mobile), DPP-Net achieves better performances: higher accuracy \& shorter inference time on these various devices. Additional experimental results also show that models found by DPP-Net achieve state-of-the-art performance on ImageNet.